\documentclass[10pt,twocolumn,letterpaper]{article}

\usepackage{iccv}

\usepackage{times}
\usepackage{epsfig}
\usepackage{graphicx}
\usepackage{grffile}
\usepackage{amsmath}
\usepackage{amssymb}
\usepackage{booktabs}
\usepackage{mwe}
\usepackage{colortbl}
\usepackage{multirow}
\usepackage[ruled,vlined]{algorithm2e}

\usepackage{xcolor}
\usepackage{listings}
\usepackage[british,american]{babel}
\usepackage{enumitem}
\usepackage{multirow}
\usepackage{textpos}  

\setlist{nosep}

\usepackage{color}

\definecolor{DarkGreen}{rgb}{0.0, 0.6, 0.2}
\definecolor{DarkerGreen}{rgb}{0.0, 0.4, 0.1}

\definecolor{DarkRed}{rgb}{0.5, 0.1, 0.1}
\definecolor{DarkerRed}{rgb}{0.5, 0.1, 0.0}

\usepackage[breaklinks=true,bookmarks=false,colorlinks=true,linkcolor=red,citecolor=blue]{hyperref}

\iccvfinalcopy 

\def\iccvPaperID{****} 

\ificcvfinal\fi

\begin{document}

\title{UVO Challenge on Video-based Open-World Segmentation 2021: 1st Place Solution \\
}

\author{Yuming Du$^{1}$ \qquad \quad Wen Guo$^{2}$ \qquad \quad Yang Xiao$^{1}$ \qquad \quad Vincent Lepetit$^{1}$
\\
$^1$LIGM, Ecole des Ponts, Univ Gustave Eiffel, CNRS, Marne-la-Vallée, France\\
$^2$Inria, Univ. Grenoble Alpes, CNRS, Grenoble INP, LJK, 38000 Grenoble, France\\
{\tt\small \{yuming.du, yang.xiao, vincent.lepetit\}@enpc.fr}, {\tt\small wen.guo@inria.fr}\\
\url{https://github.com/dulucas/UVO_Challenge}
}

\maketitle

\begin{abstract}
In this report, we introduce our (pretty straightforard) two-step ``detect-then-match'' video instance segmentation method. The first step performs instance segmentation for each frame to get a large number of instance mask proposals. The second step is to do inter-frame instance mask matching with the help of optical flow. We demonstrate that with high quality mask proposals, a simple matching mechanism is good enough for tracking. Our approach achieves the first place in the UVO 2021 Video-based Open-World Segmentation Challenge.

\end{abstract}

\section{Method}
\label{sec:method}
In this section, we present our detect-then-matching method for open-world video instance segmentation. Our method consists of two steps, the first step is to generate mask proposals for each frame of the video. The second step aims to link the detected masks using optical flow predicted by an optical flow estimator.

\subsection{First Step: Instance Segmentation}
In this section, we introduce our method for instance segmentation. We adopt a detect-then-segment pipeline. We first train an object detector to generate bounding boxes for each frame of the video. Then, we take the top-100 bounding box proposals, crop the images with these bounding boxes and feed the resized obtained image patches to a foreground/background segmentation network to obtain instance masks. We briefly introduce the networks that we used for instance segmentation, please refer to our technical report for Image-based open-world segmentation for more details\cite{du20211st}.

\begin{figure*}
\begin{center}
  \begin{tabular}{cccc}
    \includegraphics[width=1.0\linewidth]{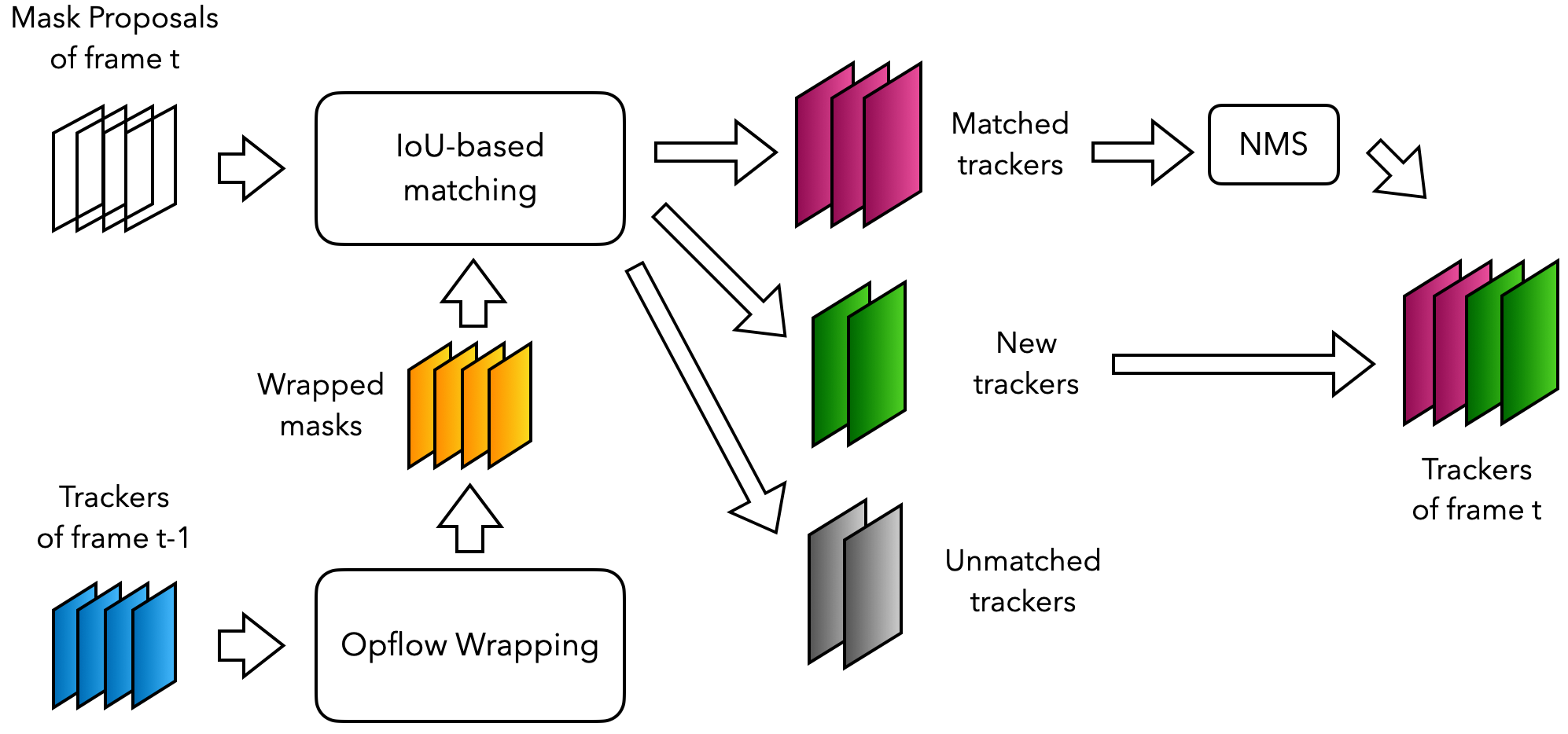}
\end{tabular}
\end{center}
\caption{\label{fig:mask matching} {\bf Overview of our mask matching pipeline.} The masks of the different trackers from frame $t-1$ are first wrapped using optical flow $F_{t-1}$, then we match the wrapped masks with detected masks from frame $t$. The trackers that do not have any matched masks will be removed if they stay without matches for 5 frames. The matched tracker will update their latest mask to their matched masks. The unmatched detected masks are used to initialize new trackers and will be added to the total trackers.}
\end{figure*}

\paragraph{Detection Network.} We adopt the Cascade Region Proposal Network~(RPN)~\cite{vu2019cascade} as our baseline network, and the Focal loss~\cite{lin2017focal} and GIoU loss~\cite{rezatofighi2019generalized} for classification and bounding box regression. We use two separate SimOTA~\cite{ge2021yolox} samplers for positive/negative example sampling during training, with one for classification and another for bounding box regression. We loose the selection criterion for bounding box regression sampler to get more positive samples for the bounding box head during training. An additional IoU branch is added in parallel with classification head and bounding box regression head to predict the IoU between the predicted bounding box and the ground truth bounding box. Following previous works~\cite{ge2021yolox, tian2019fcos}, we adopt the decoupled head to ease the conflict between classification task and regression task in object detection. Heads across all pyramid levels share the same weights to save memory. The first convolutional layer of the decoupled heads is replaced by deformable convolutional layers~\cite{dai2017deformable}. We add CARAFE~\cite{wang2019carafe} blocks in our FPN~\cite{lin2017feature} and use Swin-L transformer~\cite{liu2021swin} as our backbone network.

\paragraph{Segmentation}
We use the bounding boxes predicted by our detection network to crop image patches and resize them to 512$\times$512. The cropped image pathches are then fed to our segmentation network to get instance masks. We adopt the Upernet~\cite{xiao2018unified} architecture and Swin-L transformer~\cite{liu2021swin} as our backbone for segmentation network. The segmentation network is a binary segmentation network, the pixels are predicted as foreground if they belong to an object, otherwise as background.

\subsection{Second Step: Mask Matching across Frames}
In this section, we introduce our matching method for video instance segmentation. An overview of our method is shown in Figure \ref{fig:mask matching}. Our idea is similar to IoU-tracker. The trackers from previous frame are wrapped to current frame using predicted optical flow, then the trackers are matched with detected mask proposals from current frame by calculating the IoU~(Intersection-over-Union) between the wrapped masks of trackers and the detected masks.

We denote $M$ as the total mask proposals for all frames, with $M_t$ represents the mask proposals for frame $t$. $T$ denotes the video length and $F$ denotes the optical flows where $F_t$ represent the optical flow between frame $t$ and frame $t+1$.

We first initialize trackers with the mask proposals from the first frame $M_0$. Then, we wrap the masks of the trackers to the second frame using optical flow $F_0$. The wrapped masks are then matched with the detected masks $M_1$ by calculating the IoU between them. We consider a matching is successful only if the IoU is larger than a fixed threshold, which is 0.5 in our case. If a tracker is matched with a detected mask, we replace the latest mask of the tracker with the matched mask. If there is no match between the tracker and the masks from $M_1$, the wrapped mask is used to update its latest mask. If the tracker has not been matched continuously for 5 frames, we remove this tracker from our tracker list. For the masks from $M_1$ that are not been matched with trackers, we initialize new trackers with these masks and add these trackers to our tracker list. We use NMS~(Non-maximum-suppression) to remove the trackers whose latest masks have an IoU larger than 0.7.

We assign each tracker with a score, which is calculated as the product between the number of frames that it has been tracked and the sum of the detection scores.

\section{Dataset}
\label{sec:dataset}
\subsection{Detection}
ImageNet 22k is used to pre-train our backbone network. We then train our detectors on the COCO dataset~\cite{lin2014microsoft}. In the end, the pre-trained detectors are fine-tuned on UVO-Sparse dataset and UVO-Dense dataset~\cite{wang2021unidentified}.

\subsection{Segmentation} ImageNet 22k~\cite{imagenet_cvpr09} is used to pre-train our backbone network. We then train our segmentation network on a combination of the OpenImage~\cite{OpenImages2}, PASCALVOC~\cite{everingham2010pascal}, and COCO~\cite{lin2014microsoft} datasets. In the end, the pre-trained segmentation networks are fine-tuned on the UVO-Sparse and UVO-Dense datasets~\cite{wang2021unidentified}.

\begin{figure*}
\begin{center}
  \begin{tabular}{cccc}
    \includegraphics[width=1.0\linewidth]{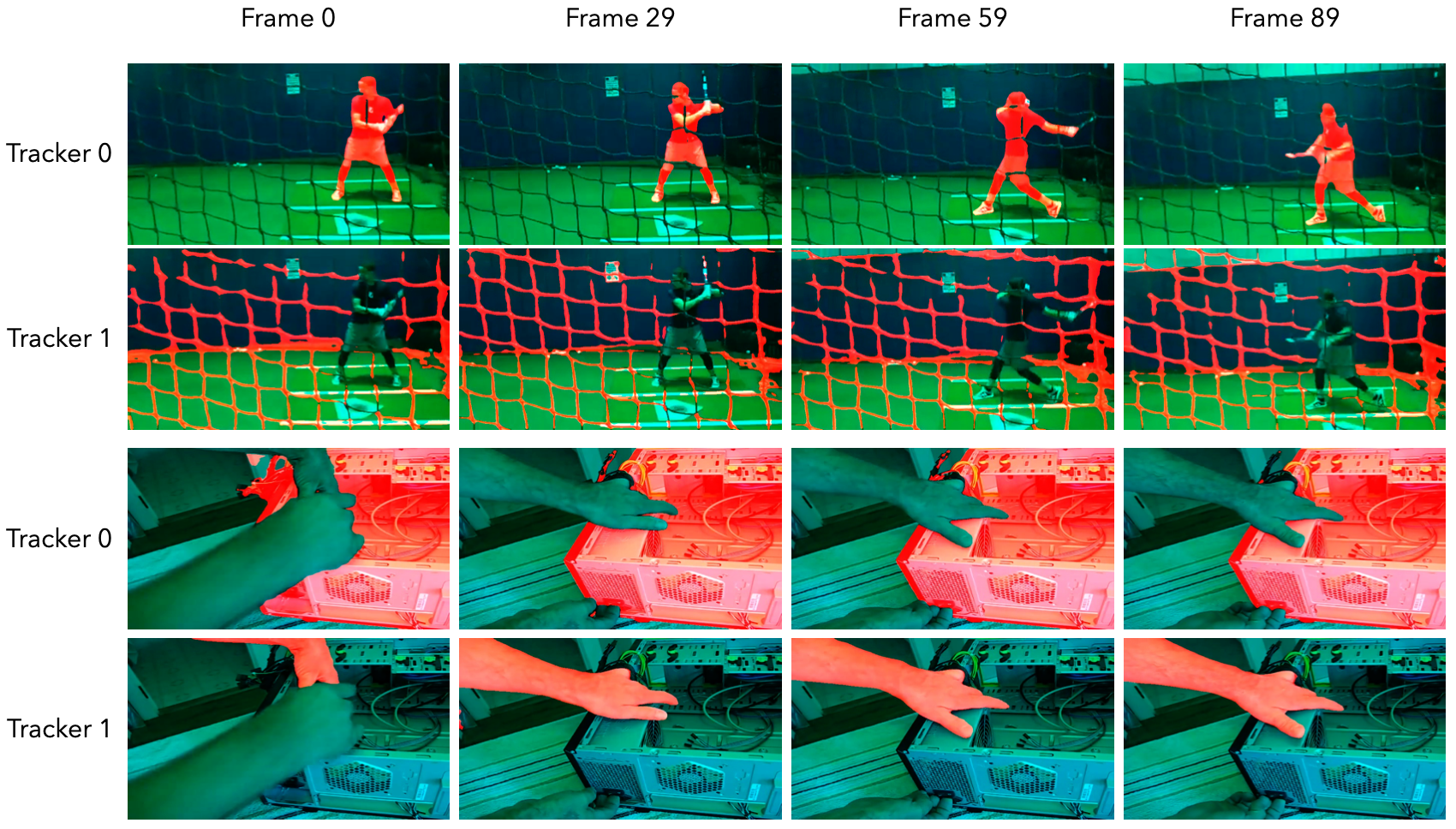}
\end{tabular}
\end{center}
\caption{\label{fig:vis challenge} Examples of results of our video instance segmentation approach on the UVO-Dense test dataset.}
\end{figure*}

\section{Implementation Details}
\label{sec:implementation details}
\subsection{Detection}
We use MMDetection~\cite{mmdetection} to train our detectors. For the backbone network, we get the Swin-L transformer pre-trained on ImageNet 22k from \footnote{\url{https://github.com/microsoft/Swin-Transformer}}. All our detectors are trained with Detectron ‘1x’ setting. For data augmentation, we use the basic data augmentation strategy as in \cite{he2017mask} for all experiments. The center ratio of both SimOTA samplers are set to 0.25, the top-K number for classification head is set to 10, while the top-K number for regression head is set to 20 to involve more positive samples. Four 3x3 conv layers are used in the classification branch and the regression head, IoU branch shares the same conv layers with the regression branch. To train the detector with Swin-L transformer backbone, we adopt the AdamW as the optimizer and set the init learning rate to 1e-4. The batch size is set to 16. After training on COCO, we fine-tune the detector on the combination of the UVO-Sparse and UVO-Dense datasets for 6 epochs. All our detectors are trained in the class-agnostic way. Test time augmentation is used during inference to further boost the network performance.

\subsection{Segmentation}
We use MMSegmentation~\cite{mmseg2020} to train our segmentation network. We use the same backbone network as our detection network. During training, given an image and an instance mask, we first generate a bounding box that envelopes the instance mask, then a 20 pixel margin is added to the bounding box in all directions. We use the generated bounding box to crop the image and resize the image patch to 512x512. Random flipping, random photometric distortion, and random bounding box jitter are used as data augmentation. We adopt 'poly' learning rate policy and set the initial learning rate to 6e-5. The batch size is set to 32 and AdamW~\cite{loshchilov2017decoupled} is used as the optimizer. We first train our network on the combination of the OpenImage~\cite{OpenImages2}, PASCALVOC~\cite{everingham2010pascal} and COCO~\cite{lin2014microsoft} datasets for 300k iterations, then we finetune the network on the combination of the UVO-Dense and UVO-Sparse datasets for 100k iterations with initial learning rate set to 6e-6. All our segmentation networks are trained in a class-agnostic way, thus, segmenting the object in the cropped path becomes a foreground/background segmentation problem. Only flip test augmentation was used during inference.

\subsection{Optical Flow Estimation} 
We use the model released by \cite{teed2020raft} trained on FlyingThings~\cite{MIFDB16}. FlyingThings is a large-scale synthetic dataset for optical flow estimation. The dataset is generated by randomizing the movement of the camera and synthetic objects collected from the ShapeNet dataset~\cite{shapenet2015}. The model for optical flow estimation is pre-trained on FlyingThings for 100k iterations with a batch size of 12, then for 100k iterations on FlyingThings3D with a batch size of 6.

\section{Visualization}
In Figure~\ref{fig:vis challenge}, we show some of our the video instance segmentation results. Our method works well for objects of different shapes.

\begin{table*}[!t]
  \addtolength{\tabcolsep}{-1.pt}
  \begin{center}
  \scalebox{1.1}
	     {
	       \begin{tabular}{@{}l | cccccc @{}}
	         \toprule
	          Teams & $AR@100$ & $AP$ & $AP@.5$ & $AP@.75$ & $AR@1$ & $AR@10$ \\ 
	         \midrule
	         
	         Baseline by host
	         & 11.77 & 7.35 & 16.33 & 6.10 & 5.06 & 11.72 \\
	         
	         CSTT(lvis\_htc\_4\_s0.002)
	         & 26.98 & 16.96 & 31.16 & 15.22 & 6.99 & 22.04 \\
	         
	         Sensetime
	         & 29.04 & 18.16 & 32.09 & 17.46 & 7.22 & 24.14 \\
	         
	         elf\_hzw (final)
	         & 34.00 & 14.13 & 23.50 & 14.52 & 7.45 & 21.14 \\
	         
	         Ours
	         & \textbf{41.17} & \textbf{27.56} & \textbf{40.61} & \textbf{29.22} & \textbf{8.96} & \textbf{29.92} \\
	         
	         \bottomrule
	       \end{tabular}
	     }
  \end{center}
  \caption{{\bf Challenge final results} on UVO-Sparse test dataset. Our method outperforms the baseline and the other submitted methods by a large margin.
  }
  \label{tab:overall ablation}
\end{table*}
\section{Potential Improvements}

Our simple ``detect-then-match'' framework can serve as a baseline for video instance segmentation. It heavily relies on the quality of mask proposals for each frame. The performance of our method could be affected by heavy occlusion, object appearance/disappearance/re-appearance, etc. These problems could be potentially solved by taking object embedding into consideration during the mask matching process.

{\small
\bibliographystyle{ieee_fullname}
\bibliography{egbib}
}

\end{document}